\documentclass[11pt,a4paper]{article}

\usepackage{latexsym}
\usepackage{amssymb}
\usepackage{amsmath}
\usepackage{amsthm}
\usepackage{float}
\usepackage{booktabs}
\usepackage{enumitem}
\usepackage{paralist}
\usepackage{multirow}
\usepackage{multicol}
\usepackage{color}
\usepackage{todonotes}
\usepackage{svg}
\usepackage{authblk}
\usepackage{hyperref}
\usepackage{comment}

\newcommand{\BibTeX}{B\kern-.05em{\sc i\kern-.025em b}\kern-.08em\TeX}

\graphicspath{{./figures/}}

\begin{document}


\title{Evaluating Large Language Models for Real-World Engineering Tasks}

\author[*]{René Heesch}
\author{Sebastian Eilermann}
\author{Alexander Windmann}
\author{Alexander Diedrich}
\author{Philipp Rosenthal}
\author{Oliver Niggemann}

\affil{Helmut Schmidt University, Germany}
\affil[*]{Corresponding author: rene.heesch@hsu-hh.de}
\maketitle


\begin{abstract}
Large Language Models (LLMs) are transformative not only for daily activities but also for engineering tasks. 
However, current evaluations of LLMs in engineering exhibit two critical shortcomings:
\textit{(i)} the reliance on simplified use cases, often adapted from examination materials where correctness is easily verifiable, and
\textit{(ii)} the use of ad hoc scenarios that insufficiently capture critical engineering competencies.
Consequently, the assessment of LLMs on complex, real-world engineering problems remains largely unexplored.
This paper addresses this gap by introducing a curated database comprising over 100 questions derived from authentic, production-oriented engineering scenarios, systematically designed to cover core competencies such as product design, prognosis, and diagnosis.
Using this dataset, we evaluate four state-of-the-art LLMs, including both cloud-based and locally hosted instances, to systematically investigate their performance on complex engineering tasks.
Our results show that LLMs demonstrate strengths in basic temporal and structural reasoning but struggle significantly with abstract reasoning, formal modeling, and context-sensitive engineering logic.
\end{abstract}


\section{Introduction}

Large Language Models (LLMs) have become increasingly prominent in engineering, demonstrating potential in areas such as requirements verification \cite{reinpold2024exploring}, industrial process control \cite{xia2024control}, and simulation model generation \cite{merkelbach2024using}.
Recently, LLM-powered AI agents have emerged to automate complex engineering workflows \cite{wangSurveyLargeLanguage2024}.
However, engineering applications pose specific challenges compared to many other fields, including requirements for reliability, interpretability, and compliance with rigorous safety and security standards \cite{Windmann24}.

Despite notable advancements, the applicability and reliability of LLMs within engineering contexts remain insufficiently understood. 
While concerns regarding interpretability are widely acknowledged across neural network-based methods, engineering-specific challenges have received comparatively less attention. 
Previously published evaluations rely on simplified use cases, often adapted from examination materials where correctness is easily verifiable, or use of ad hoc scenarios that insufficiently capture critical engineering competencies \cite{suresh2024gpt}.
Particularly critical are the capabilities of LLMs in spatial reasoning, inference of implicit design intentions, reliability of generated models, prediction of system behaviors, and temporal-causal reasoning---all capabilities typical for engineering.

In the following, specific research questions (RQ) are derived.
Engineering systems must be reliable, exhibiting predictable behavior, minimizing failures and downtime, with rigorous methods such as failure analysis, uncertainty modeling, and redundancy ensuring robustness. 
Such methods rely on the availability of accurate and consistent engineering models---world models in AI terms. Traditionally, these are created manually by expert knowledge. 

\textit{\textbf{RQ1:} Can LLMs generate sufficiently correct and consistent world models to meet the reliability standards required in engineering applications?}

\smallskip
Further, engineering models often exhibit hierarchical, graph-structured representations capturing modularity, object interactions, connectivity, and flows of energy, information, or materials. 
These structures frequently encode explicit spatial relationships (e.g., Euclidean layouts in industrial plants), prompting the second RQ:

\textit{\textbf{RQ2:} Can LLMs accurately interpret and manage transitional and non-local relations between engineering entities?}

\smallskip
Moreover, engineering is inherently concerned with the design of technical systems, wherein models embed implicit design intentions and constraints. I.e., a technical specification can only be understood, and optimized, against the background of the later intended use. For example, the design of a production facility typically aims at achieving specific production goals while satisfying constraints on quality, energy efficiency, cost-effectiveness, and workforce capabilities.

\textit{\textbf{RQ3:} How effectively can LLMs infer implicit design intentions underlying engineering models and product architectures?}

\smallskip
Additionally, engineering systems, when dealing with physical entities, are intrinsically temporal and causal \cite{vukovic2022causal}. 
I.e., technical processes exhibit dynamic sequences where events causally influence subsequent states, ranging from simple correlations to complex, conditionally dependent chains. This means, changes at some of the system's input may lead to significant deviations in a product's specifications or the system's functionality. It is therefore possible to pose the fourth research question.

\textit{\textbf{RQ4:} Are LLMs capable of reasoning about temporal event sequences and capturing complex causal dependencies over extended event chains?}

\smallskip
Further, predictive capability is essential for engineering analysis and simulation. In engineering, behavior models are used in the design and in the operation phase to assess system qualities. Here, predicting system behavior involves managing dependent, high-dimensional state variables and their changes over time, often under non-linear and feedback-rich conditions.

\textit{\textbf{RQ5:} How accurately can LLMs predict local system behaviors, such as time series data, and what are their limitations when faced with non-linear, complex behavior?}

\smallskip
To address these research questions, four state-of-the-art LLMs were systematically evaluated using two distinct task types. 
For RQ1, RQ2, and RQ4, the models were presented with a curated set of over 80 new questions derived from both a real-world production system and a production simulation model originally designed for AI validation. For RQ3, a separate set of 20 domain-specific questions focused on product design was used to assess the ability of LLMs to infer design intentions from function-based representations.
Each question was constructed to target a specific, measurable aspect corresponding to one of the defined research questions.
To evaluate forecasting capabilities (RQ5), the LLMs were additionally tasked with predicting future steps based on a sequence of previously observed steps from 200 sample time series.
To the best of our knowledge, this is one of the first contributions using such a broad range of engineering tasks and such complex examples. 

Accounting for the concerns about the unauthorized use of data without the company’s consent when AI models are operated in the cloud \cite{Windmann24}, three locally hosted LLMs were employed.
Additionally, to assess whether higher performance could be achieved by leveraging a cloud-based model without privacy constraints, GPT-4o \cite{hurst2024gpt} was included under identical conditions.

The contribution of this paper is two-fold. 
First, it introduces a new (publicly available) curated dataset of over 100 domain-specific questions as well as forecasting tasks reflecting real-world engineering challenges, facilitating systematic evaluation of LLM capabilities. These questions are derived from a systematic analysis of engineering capabilities and by using two new complex system models.
Second, it provides a comprehensive analysis of the strengths and limitations of state-of-the-art LLMs in representing, reasoning with, and applying engineering knowledge.

\section{State of the Art}

LLMs have demonstrated strong performance across general-purpose tasks involving factual recall, commonsense reasoning, and linguistic coherence. 
However, their applicability to domain-specific engineering challenges, such as structured model interpretation, implicit constraint handling, fault diagnosis, and predictive analysis of system behavior remains underexplored and insufficiently validated.

A growing body of research has examined the extent to which LLMs construct internal world models, i.e. consistent latent representations of the situations or systems under discussion. 
Several studies suggest that very large LLMs are capable of maintaining such models, supporting coherent inferences and scenario reasoning \cite{feng2024monitoringlatentworldstates, li2024emergentworldrepresentationsexploring}. 
These internal representations primarily serve reasoning tasks, with evidence that LLMs can exhibit strong logical and commonsense reasoning when sufficiently scaled \cite{huang2023reasoninglargelanguagemodels, plaat2024reasoninglargelanguagemodels}.
Further investigations have probed whether these world models are grounded in factual knowledge \cite{patel2022mapping}, or whether they align with cognitive models of human reasoning \cite{Chuganskaya.2023}. 
In a complementary line of inquiry, \cite{xu2025large} examined LLMs’ reasoning abilities within the formal logic frameworks foundational to engineering and computer science. 
Similarly, the capacity of LLMs to model causal relationships was investigated by \cite{ban2025llm}. 
The extent to which LLMs can represent and reason over structural dependencies remains an area of active research. 
Recent advancements suggest that LLMs are proficient in interpreting local and well-defined relational structures, particularly those found in graph-based and Euclidean spatial domains. 
Nevertheless, their capacity to model more intricate, non-local, and dynamic relational forms is less well understood \cite{yamada2023evaluating,zhang2024llm4dyg}.

In parallel, several works have explored the application of LLMs in concrete engineering contexts, particularly in the domain of cyber-physical systems (CPS). 
\cite{hirtreiter2024toward} investigated the automated synthesis of control structures using LLMs, while \cite{ogundare2023resiliency} analyzed the robustness of LLM-generated system models. 
However, the latter’s models were limited to isolated equations without interdependencies, restricting their utility for system-level reasoning.
Kato et al. \cite{kato2023simple} focused on extracting equations from scientific corpora and assembling them into physical models, leveraging LLMs to assess equation equivalence and enforce domain-specific constraints. 
In a different engineering application, \cite{peifeng2024joint} employed LLMs for fault diagnosis in aviation assembly. 
However, the approach remained largely constrained to text extraction and did not incorporate fault reasoning. 
Similarly, \cite{saba2023text} proposed a text-based solution for technical documentation analysis but did not extend it to reasoning or diagnosis.
\cite{vicente2022gutenbrain} proposed the use of metadata and question-answering systems for knowledge extraction from technical manuals, but their system lacked integration with structured diagnostic frameworks. 
Knowledge mining approaches, such as those by \cite{meier_knowledge_2024}, and software fault localization efforts by \cite{kang2023preliminary} and \cite{wu2023large}, offer insights into LLM-based reasoning in software engineering, though these remain narrowly scoped. 
Likewise, \cite{balhorn2024toward} addressed the correction of P\&ID diagrams using LLMs, yet did not extend their methodology to model generation or analysis.
An alternative paradigm was introduced by \cite{sowa2025expert} through LLMs, fine-tuned for high performance in narrow domains, called Generative Artificial Expert. 
While promising, such domain-specialized models raise concerns regarding scalability and generalizability to multi-domain engineering scenarios.

In summary, while individual studies have explored aspects of LLM application in engineering, there remains a lack of systematic, cross-domain evaluation of their capabilities and limitations. 
Specifically, there is a need for a structured and comprehensive analysis that aligns with the unique requirements of engineering disciplines, beyond narrow or illustrative application cases.

\section{Analysis of Research Questions}

Given the confidential nature of the detailed process and plant information relevant to the engineering tasks, the evaluation of the research questions mainly focused on locally-deployed instances of LLMs.
The local LLM instances were deployed utilizing the Ollama\footnote{\url{https://ollama.com/}} and the vLLM\footnote{\url{https://docs.vllm.ai/en/stable/}} framework, hosted on two workstations that are equipped with three NVIDIA L40 and two A100 graphics processing units (GPUs), respectively.
Additionally, to account for the significant higher performance of cloud hosted LLMs, an additional model instance was accessed via an online service.
The four LLMs evaluated in this study were: Qwen2.5 \cite{yangQwen2TechnicalReport2024}\footnote{\url{https://huggingface.co/Qwen/Qwen2.5-72B-Instruct}}, DeepSeek-R1 (DS-R1) \cite{deepseek-aiDeepSeekR1IncentivizingReasoning2025}\footnote{\url{https://huggingface.co/deepseek-ai/DeepSeek-R1-Distill-Llama-70B}}, LLaMA 3.3 \cite{grattafioriLlama3Herd2024}\footnote{\url{https://huggingface.co/meta-llama/Llama-3.3-70B-Instruct}} and GPT-4o \cite{hurst2024gpt}.

For the first four research questions, a corresponding set of questions was formulated and systematically categorized according to distinct, measurable capability domains.
To account for stochastic variability inherent in LLM outputs, each question was posed multiple times \cite{vranjevs2024design}. 
Recognizing the critical importance of reliability in engineering contexts, only the least accurate response for each question was considered for evaluation. 
The responses were assessed using a three-point scale: a score of 0 indicated an incorrect or nonsensical response; a score of 0.5 reflected a partially correct response, consistent with reasoning skills expected from a junior engineer; and a score of 1 denoted a comprehensive and accurate analysis, characteristic of the expertise of a senior engineer.
Finally, the scores within each category were averaged.

For RQ1, RQ2, and RQ4, both a real industrial plant and a simulated production environment served as the primary use cases.

\begin{figure}
    \centering
    \includegraphics[scale=0.45]{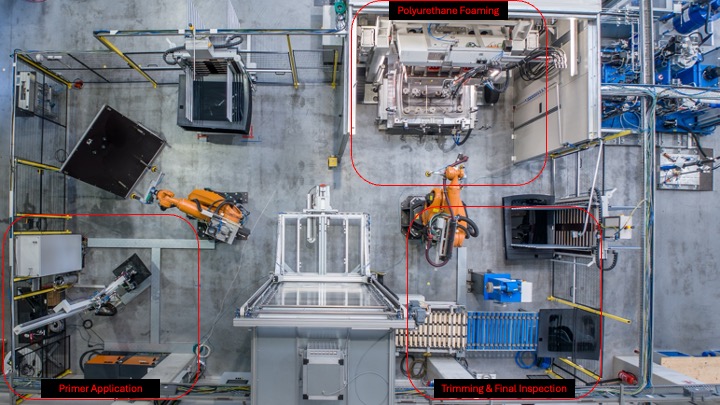}
    \caption{Modular glass foaming plant that served as first use case. The productions sells \textit{Primer Application}, \textit{Polyurethane Foaming}, and \textit{Trimming \& Final Inspection} are highlighted.}
    \label{fig:foaming_plant}
\end{figure}

\noindent \emph{\textbf{Real Plant:}} 
The first system is a physical glass foaming plant (cf. Figure \ref{fig:foaming_plant}), selected as a representative example of a highly structured, multiphase industrial process. 
The plant comprises three fully automated, interdependent production cells: primer application, polyurethane foaming, and trimming with final inspection. 
Each stage integrates specialized machinery, robotic handling, material transformation processes, and real-time quality control mechanisms.
The production flow necessitates conditional decision-making (e.g., based on quality control outcomes), temporally constrained operations (e.g., curing and evaporation times), and spatial dependencies (e.g., inter-cell material transfers). 
The plant’s modular architecture and iterative complexity, coupled with strict requirements for internal consistency, impose significant cognitive demands on LLMs. 
Specifically, they must reason across sequential operations, track spatial-temporal system states, and adapt to dynamically evolving configurations. 
Consequently, this setting offers an industrially grounded, scalable platform to rigorously evaluate the ability of LLMs.

\noindent \emph{\textbf{Simulated Plant:}} 
The second system is a simulated discrete manufacturing environment that models an end-to-end production process across seven modular stages: Incoming Goods, Inspection, Sorting, Storage, Processing, Packaging, and Outgoing Goods. 
It features detailed, parameterizable representations of various equipment, including conveyors, portal robots, autonomous mobile robots (AMRs), and six-axis robots, each exhibiting distinct operational parameters and failure modes.
The system’s structural complexity, dynamic material flow, and tightly coupled interdependencies provide a comprehensive context for evaluating LLM performance in industrial engineering tasks. 
These tasks include mechanical operations reasoning, maintaining data integrity, diagnosing system faults, interpreting control logic, and optimizing performance within a data-intensive and operationally realistic framework.

For RQ5 regarding the forecasting capability, the setup was slightly different.
Each model was tasked with forecasting 10 future steps based on 30 previous steps.
The forecasting accuracy was evaluated using Mean Squared Error (MSE) on 200 samples from two standardized CPS datasets.
The prompting consisted of a general task description, a dataset-specific description, and input time-series data in CSV format.
Multiple prompting strategies have been tested for each model and dataset.

The detailed results including the rated answers and the corresponding scores are available in the GitHub repository\footnote{\url{https://github.com/imb-hsu/Wrenchmark}}.

\subsection{RQ1: Can LLMs generate world models?}

Human reasoning is based on (more or less) consistent world models. Whenever we perceive and act in the world, we create a world model which allows us to reason, e.g., to predict possible future outcomes of actions. Such world models trade correctness for consistency, i.e., they predict exactly one outcome in each situation while this outcome might be (partially) incorrect. 
Imprecisions (i.e. incorrectness) are acceptable as long as predictions can be made quickly and help to optimize decisions. E.g. for calculating trajectories, Newtonian physics is sufficient, even though physically superior models such as the theory of relativity are known. Humans overcome limitations of world models normally simply by having several world models, e.g. physicists switch between Newtonian physics and Einstein physics easily. While each world model is consistent, no consistency is required between world models, i.e., for each type of decision we tend to rely on one specific world model.

Models are also at the heart of all engineering processes. And again, people use different models for different tasks, e.g., models at early design phases differ from models at operation time. These models again predict consequences of actions. Engineers normally create such models manually, a good way of looking at these models is as a reflections of an engineers mental world model. So the questions arises whether LLMs also create such internal world models. 

\smallskip
What are measurable features of world models which can here be used to verify whether LLMs use internally world models?

\noindent \emph{Iterative Consistency:\ }
World models are consistent even if new information is incrementally added. This new information might extend the original model spatially or might add a new aspect. While it is straightforward to keep consecutive modifications consistent, it is rather difficult to guarantee consistency over a large number of steps.

Which experimental design can analyze this feature? Several production systems are built incrementally. The increments are randomly, so it is hard to keep track of the overall system. Later, small changes are applied to the system. After each change, it is checked whether the overall system behavior is still predicted correctly. 

\noindent \emph{Closed-World Assumption:\ }
The closed-world assumption states that everything which cannot be deduced from the model is not true. E.g., if a model does not predict a failure, this failure has not occurred. Please note that this assumes that the model captures all relevant system behavior. 

Which experimental design can analyze this feature? Again, several productions systems are used. This feature is analyzed by checking whether not-modeled aspects are predicted correctly.

\noindent \emph{Non-Locality of Consistency:\ }
Technical systems consist of a network of interconnected modules where each module might have similar substructures. Modules are connected via energy, material, products or information. The effect of one module onto another module is easily analyzed if they are connected directly, indirect and transitional effects are much harder to predict.

Which experimental design can analyze this feature? The two aforementioned production plants are used and transitional effects are analyzed.

\noindent \emph{Concept Generation on Type-Level:\ }
LLMs learn concepts within their latent layers. For engineering tasks, learning concepts on a instance-level, i.e., system-specific concepts such as a special machine, are rather meaningless since these information are often provided as an input. But learning concepts on a type-level, i.e., abstract, reusable concepts such as robots or conveyors, is helpful since it allows to reuse results from one system for other systems and such type-level concepts are often used to identity optimization points in the system, e.g., to replace conveyors with autonomous vehicles.

This research question is empirically examined using 45 questions, distributed across the four measurable capabilities. 

The results of the evaluation are summarized in Table~\ref{tab:res_rq1}.
\textit{Iterative Consistency} was comparatively tractable, with most models maintaining stable internal representations under incremental updates. 
Similarly, strong performance under the \textit{Closed-World Assumption} suggests that LLMs can constrain inferences to explicitly modeled system elements when representations are sufficiently complete. 
However, reasoning extends local scope seems to remain challenging. 
\textit{Non-Locality of Consistency} yielded moderate results, highlighting limitations in capturing cross-component dependencies and sustaining coherence across modular structures. 
The biggest weakness was in \textit{Concept Generation on Type-Level}, where models failed to abstract reusable, system-independent knowledge from instance-level inputs. 
As such abstractions are central to engineering tasks like transfer, optimization, and modular design, these findings indicate that while LLMs can model localized system behavior, their current capacity to construct reliable, generalizable world models do not meet the requirements of engineering tasks.
Overall, the largest LLM (GPT-4o) consistantly performed better than the smaller, locally hosted models.

\begin{table}[H]
    \centering  
    \caption{The table summarizes the analysis of LLMs’ capabilities to generate a consistent world model for engineering tasks. The score shown corresponds to the average of all questions within the respective category, on a scale from 0 to 1.}
    \begin{tabular}{lcccc}  
    \toprule 
    \textbf{Feature} & \textbf{Qwen2.5} & \textbf{DS-R1} & \textbf{LLaMa 3.3} & \textbf{GPT-4o}\\ 
    \midrule
    Iterative Consistency        & 0.78 & 0.63 & 0.69 & \textbf{0.91} \\
    Closed World                 & 0.61 & 0.72 & \textbf{0.94} & \textbf{0.94} \\
    Non-Locality                 & 0.65 & 0.70 & 0.65 & \textbf{0.95} \\
    Type-Level                   & 0.40 & 0.55 & 0.70 & \textbf{0.85} \\
    \bottomrule
    \end{tabular}
    \label{tab:res_rq1}
\end{table}

\subsection{RQ2: Can LLMs accurately interpret and manage transitional and non-local relations?}
A defining characteristic of engineering problems lies in their inherently spatial and relational structure. Such problems frequently emerge from complex systems comprising interconnected components that operate across multiple domains, e.g., mechanical, electrical, thermal, and informational domains. 
These systems are not simply aggregates of isolated modules; rather, they exhibit dependencies that span across subsystems, hierarchical layers, and temporal stages of operation or design \cite{elmaraghy2012complexity,efatmaneshnik2016complexity,kocher2022research}.
Consequently, the ability to reason about transitional (i.e., sequential, conditional, or time-dependent) and non-local (i.e., cross-component, cross-layer, or system-wide) relationships constitutes a core requirement in engineering.
To evaluate the ability of LLMs to engage with complex and transitional relationships in engineering contexts, this subsection focuses on five measurable capabilities:

\noindent \emph{Causal Inference Ability:\ } The capacity of the model to infer causal relationships between events or system changes.
Given a system fault, can the LLM infer the preceding chain of state transitions that led to the failure?

\noindent \emph{State Transition Comprehension:\ } The ability to track and describe the evolution of a system across multiple states.
Given a multistep mechanical or electrical process, can the model accurately predict or articulate intermediate and final states?

\noindent \emph{Multi-Variable Dependency Resolution:\ } The ability to manage interdependent variables within a system (e.g., temperature, pressure, load).
Given an adjustment to a single parameter, can the model accurately predict the system’s response under established constraints?

\noindent \emph{Modularity and System Integration Reasoning:\ } The capacity to understand interactions among subsystems and the implications of localized changes.
Given a substitution of a subsystem, can the model infer the resulting effects on system-level behavior?

\noindent \emph{Sequential Understanding:\ } The ability to reason over temporally ordered operations or events.
Given a partial action sequence and a target system state, can the model identify the next correct step, or identify inconsistencies in the execution order? 

This research question is empirically examined using 20 questions, equally distributed across the five measurable capabilities, within the context of the two aforementioned production systems. 
The results are summarized in Table \ref{tab:res_rq2}. 
LLMs exhibited good performance in interpreting transitional relations, particularly in tasks involving \textit{State Transition Comprehension} and \textit{Sequential Understanding}. This suggests that reasoning over temporally structured processes and tracking discrete system states fall within the operational grasp of current LLMs. 
However, when the reasoning task extends to non-local relations, performance markedly decreases. 
\textit{Multi-Variable Dependency Resolution} proved especially challenging, with uniformly low scores across models, revealing a substantial limitation in capturing interdependent relationships that span multiple components or layers of abstraction. 
Similarly, while \textit{Causal Inference} yielded moderate performance, the results reflect ongoing difficulty in tracing causal mechanisms and propagating effects beyond immediate local context. 
Lastly, performance on \textit{Modularity and System Integration Reasoning} was inconsistent, indicating that while LLMs can occasionally extrapolate local changes to system-wide consequences, this capacity is unreliable and sensitive to contextual variability. 
Overall, the findings suggest that while transitional reasoning is partially within reach, robust handling of non-local relational structures remains an open challenge for current LLM architectures.

\begin{table}[H]
    \caption{The table summarizes the analysis of LLMs’ capabilities in handling spatial structures. The score shown corresponds to the average of all questions within the respective category, on a scale from 0 to 1.}
    \centering
    \begin{tabular}{lcccc}
    \toprule
    \textbf{Feature} & \textbf{Qwen2.5} & \textbf{DS-R1} & \textbf{LLaMa 3.3} & \textbf{GPT-4o}\\ 
    \midrule
    Causal Inference                    & \textbf{0.63} & 0.50 & 0.50 & 0.50 \\
    State Transition                    & \textbf{1.00} & \textbf{1.00} & 0.63 & \textbf{1.00} \\
    Multi-Var. Dep.              & 0.38 & \textbf{0.63} & 0.38 & 0.38 \\
    Modularity                          & 0.75 & 0.63 & \textbf{0.88}& \textbf{0.88 }\\
    Seq. Understanding                  & 0.75 & 0.88 & 0.63 & \textbf{1.00} \\
    \bottomrule
    \end{tabular}
    \label{tab:res_rq2}
\end{table}

\subsection{RQ3: How effectively can LLMs infer implicit design intentions underlying engineering models?}

Engineering models and product architectures are not just collections of technical components—they embody a purpose. The structure of a system reflects trade-offs between performance, cost, safety, sustainability, and other often conflicting goals. In early product design, especially during conceptual phases, design intentions are frequently not documented explicitly but are embedded within function structures, flow models, and system architectures~\cite{Pahl.2007, Stone.1999}. Human engineers interpret these structures by drawing on domain knowledge and contextual understanding of design goals.

In engineering, a system's intention is crucial to evaluate design decisions, validating functionality, or proposing optimizations. Studies in design cognition emphasize that intention plays a central role in how engineers interpret, modify, and communicate design artifacts~\cite{Crilly.2010}. However, LLMs tend to focus on syntactic or surface-level associations and often fail to recognize design intentions or rationale, even when functional models or complete system descriptions are provided~\cite{Chuganskaya.2023, xu2025large}. This section investigates whether LLMs are capable of identifying design intentions from structural input alone, and whether they can distinguish meaningful, goal-directed engineering decisions from superficial or absurd variations.

To empirically evaluate this research question, we designed a series of test questions aimed at assessing whether state-of-the-art LLMs can infer design intentions from partial or structured engineering information. These questions focus on tasks where goals, constraints, and purpose are not made explicit but are implicitly embedded in functional architectures, process sequences, or design decisions. The central question is whether LLMs are capable of interpreting not just what a system does, but why it is structured in a particular way.

The evaluation is structured into five distinct categories, each reflecting a specific aspect of intention inference in engineering design.

\noindent \textit{Inferring Product Purpose:\ } Tests whether the LLM can identify the intended output or application of a system based on its functional structure or component interactions.

\noindent \textit{Evaluating Use-Case Fit:\ } Asks whether the LLM can assess whether a given system is suitable for producing a specified product or fulfilling a particular function.

\noindent \textit{Optimization Trade-offs:\ } Probes whether the LLM is able to consider implicit constraints such as cost, energy, material usage, or feasibility when evaluating design decisions.

\noindent \textit{Rationale Behind Design Choices:\ } Tests the model’s ability to explain the reasoning behind specific architectural decisions, such as the sequencing of functions or the inclusion of certain elements.

\noindent \textit{Recognizing Design Absurdities:\ } Examines whether the LLM can reject obviously inappropriate or illogical suggestions by reasoning about the underlying design logic.

\begin{table}[H]
    \centering
    \caption{The table summarizes the analysis of LLMs’ capabilities to infer design intentions. The score shown corresponds to the average of all questions within the respective category, on a scale from 0 to 1.}
    \begin{tabular}{lcccc}
    \toprule
    \textbf{Feature} & \textbf{Qwen2.5} & \textbf{DS-R1} & \textbf{LLaMa 3.3} & \textbf{GPT-4o}\\ 
    \midrule
    Product Purpose        & 0.50 & 0.38 & \textbf{0.63} & \textbf{0.63} \\
    Use-Case Fit                 & 0.63 & 0.63 & 0.50 & \textbf{0.88} \\
    Optimization                 & 0.50 & 0.38 & 0.25 & \textbf{0.63} \\
    Rationale                   & 0.63 & 0.50 & 0.50 & \textbf{0.88} \\
    Absurdities                   & 0.25 & \textbf{0.63} & 0.25 & 0.38 \\
    \bottomrule
    \end{tabular}
    \label{tab:res_rq3}
\end{table}

The results are shown in Table~\ref{tab:res_rq3}. In tasks centered on \textit{Product Purpose} and \textit{Use-Case Fit}, models showed a moderate but noticeable ability to associate abstract function structures with real-world products. In many cases, models inferred correct or plausible application domains even in the absence of explicit labels, suggesting that LLMs can, to some extent, reconstruct high-level intent from structural patterns alone. However, performance declined significantly when intentionality became more implicit or multidimensional. In the category of \textit{Optimization Trade-offs}, models struggled to weigh competing constraints such as cost, performance, or feasibility—especially when these trade-offs were context-sensitive. Their responses often defaulted to generic reasoning or failed to account for diminishing returns and design constraints that are second nature to human engineers. The most difficult category was \textit{Recognizing Design Absurdities}. While some questions intentionally embedded implausible or excessive configurations (e.g., overpowered motors or invalid function structures), others relied on subtler cues of contextual mismatch. Here, most LLMs failed to challenge technically valid but inappropriate solutions. This reveals a limitation in their ability to internalize engineering norms and distinguish between functional correctness and contextual relevance. Interestingly, even when LLMs successfully identified correct mappings or produced coherent reasoning, their justifications often lacked depth or overlooked critical technical details. This was particularly evident in the \textit{Design Rationale} category, where questions required reflecting on why certain functions were positioned or combined in specific ways. While the surface-level logic was often addressed, deeper insights about modularity, abstraction, or constraint propagation were rarely articulated with clarity. Overall, the results highlight a distinct gap between syntactic and semantic understanding. LLMs can often follow structural logic, but fall short in recognizing embedded intent, evaluating trade-offs, or reasoning about design fitness beyond surface-level patterns.

\subsection{RQ4: Are LLMs capable of temporal reasoning and capturing complex causal dependencies?}
Causal reasoning, either temporal or spatial, was for a long time a purely human trait. With experiments such as determining object permanence \cite{fields2021object} researchers studied whether infants and animals are able to track the potential location of objects, although they may be temporarily hidden from their perception. Related experiments are executed in counterfactual reasoning \cite{pearl2009causality}. Questions about causal relationships were also tackled by the diagnostic research community, which attempts to emulate human reasoning to diagnose technical systems; even today, many of those questions remain unanswered for technical systems \cite{pill2024challenges}. However, there is reason to believe that LLMs may be able to perform temporal, spatial, and counterfactual reasoning in technical systems that may be on par with human reasoning.

To answer this research question, we look at four major categories of reasoning. 

\noindent \textit{Logical Diagnosis:\ } The LLM's ability to perform standard fault diagnosis \cite{feldman2010approximate}. We look at a task to create a logical model and perform fault diagnosis, including the generation of Python source code. We determine whether the LLM is able to distinguish simple backward-reasoning tasks: Something happened, what might have been the cause? This involves understanding the input data and being able to reason from some conclusion toward possible causes. 

\noindent \textit{Detecting Causalities:\ } The LLM's ability to detect causal relationships and performing forward and backward reasoning \cite{pearl2009causality, bochman2021logical}. Here we investigate the LLM's ability to create a conceptual causal model from an unknown problem description. These problems were created only for the present article. For some experiments, we used the glass production scenario introduced above. 

\noindent \textit{Logical Modeling:\ } The LLM's ability to understand and use strong-fault propositional logic models (i.e., models with known failure behavior) \cite{struss1989physical}. Most diagnosis literature has focused on weak-fault models (i.e., models that describe only a system's normal behavior). Here the LLM has to formulate strong-fault models and combine this knowledge with known causal dependencies. 

\noindent \textit{Physical Modeling:\ } The LLMs capabilities to create and reason about physical models \cite{frisk2017toolbox}. In contrast to logical models, physical models depend on causalities, physical parameters, and real numbers (as compared to Boolean values) that the LLM has to deduce and connect correctly. This is a complication both for the model, as for the reasoning with the model.

The results of the evaluation of this research question are summarized in Table \ref{tab:res_rq4}.
Among the assessed capabilities, \textit{Logical Diagnosis}, a well understood domain with a significant amount of literature, proved to be easy for most of the evaluated LLMs, which is a trivial result given the large amount of literature on the topic.
Similarly, \textit{Detecting Causalities} from textual descriptions, while more abstract, showed some success, indicating that LLMs can often infer simple causal chains when the relationships are presented in narrative form. 
However, constructing a coherent and formal causal model while respecting known formalisms still appears to exceed current capabilities.
In particular, \textit{Logical Modeling} with strong-fault models introduces a notable increase in task complexity. 
Here, not only syntactic correctness is required, but also elaborate semantic integration of causal structures, fault modes, and logical dependencies. 
Most models failed to generate valid propositional models despite producing fluent natural language explanations, revealing a disconnect between narrative reasoning and formal model construction. 
\textit{Physical Modeling}, which requires integration of numerical reasoning, causal dependencies, and physical constraints, also proved to be difficult. 
With minor exceptions, all models exhibited severe limitations, with either hallucinating outputs or failing to create physically meaningful relationships. 
Overall, the results indicate that while current LLMs are competent at identifying and reasoning over simple temporal and causal relations and tasks close to their training data, they are ill-equipped to engage with more sophisticated forms of causal modeling that require formalisms and domain-specific structure.

\begin{table}[H]
    \centering
    \caption{The table summarizes the analysis of LLMs’ capabilities in handling modeling for fault diagnosis. The score shown corresponds to the average of all questions within the respective category, on a scale from 0 to 1.}
    \begin{tabular}{lcccc}
    \toprule
    \textbf{Feature} & \textbf{Qwen2.5} & \textbf{DS-R1} & \textbf{LLaMa 3.3} & \textbf{GPT-4o}\\ 
    \midrule
    Logical Diagnosis      & \textbf{1.00} & 0.50 & 0.50 & \textbf{1.00} \\
    Detecting Causalities  & 0.63 & \textbf{1.00} & 0.88 & 0.75 \\
    Logical Modeling      & 0.32 & 0.33 & 0.00 & \textbf{1.00}\\
    Physical Modeling     & \textbf{0.50} & 0.13 & \textbf{0.50} & \textbf{0.50} \\
    \bottomrule
    \end{tabular}
    \label{tab:res_rq4}
\end{table}

\subsection{RQ5: How accurately can LLMs predict system behavior?}

Accurate forecasting in CPS remains challenging due to their inherent dynamic complexity, yet it is essential for tasks like prognostics and health management. Recently, general-purpose LLMs have attracted attention as potential forecasting tools, hypothesizing that their broad understanding from pretraining may enable effective predictions even without fine-tuning, if prompted appropriately \cite{tangTimeSeriesForecasting2025}.

To investigate this capability, we evaluate forecasting performance on two datasets representing distinct CPS characteristics:
\begin{compactitem}
\item \textit{Three-Tank} \cite{steudeLearningPhysicalConcepts2022}: Simulated interconnected tanks exhibiting simple, cyclical dynamics.
\item \textit{ETTh1} \cite{zhouInformerEfficientTransformer2021}: Real-world hourly transformer data with complex daily and seasonal patterns.
\end{compactitem}

\noindent
The LLMs were compared against two specialized baseline models: LSTM \cite{hochreiterLongShortTermMemory1997} and DLinear \cite{zengAreTransformersEffective2023}.
These baseline models were selected due to their proven forecasting efficiency despite significantly smaller parameter counts of 1M and 5k, respectively.
Both models underwent hyperparameter tuning via grid search, while the LLMs were tested in a zero-shot setting with controlled prompting (temperature = 0).
To assess whether LLMs truly comprehend the underlying CPS, various prompting strategies were examined.


The results are summarized in Table~\ref{tab:RQ5_prompt_strategies}.
The specialized baseline models clearly outperform general-purpose LLMs, despite their significant lower parameter count.
Qwen2.5 and LLaMA 3.3 performed relatively well on the simpler Three-Tank dataset, indicating a capacity to capture basic cyclic patterns. 
However, performance degraded substantially on the more complex ETTh1 dataset, where no LLM performed satisfactory. 
As expected, extending prompts with explicit system equations for the Three-Tank dataset improved the forecasting performance for most models.
However, performance on the more complex ETTh1 dataset does not improve when given an extensive description of the underlying physical system.
In fact, adding no or even the wrong dataset description can actually improve model performance, suggesting the models benefit primarily from capturing general cyclic patterns rather than true system understanding.
However, the complete absence of instructions consistently worsened performance, highlighting the necessity of basic task framing.
Overall, general-purpose LLMs demonstrate only limited forecasting accuracy in complex real-world CPS.


\begin{table}[H]
\centering
\caption{Forecasting Performance Comparison (MSE). For each prompting strategy, the best performance per model is highlighted in bold.}
\begin{tabular}{llcc}
\toprule
\textbf{Model} & \textbf{Prompt Strategy} & \textbf{Three-Tank} & \textbf{ETTh1} \\
\midrule
LSTM & - & 0.0063 & 0.7009 \\
\midrule
DLinear & - & 0.0100 & 0.4162 \\
\midrule
\multirow{5}{*}{Qwen2.5} & Concise & 0.0116 & \textbf{1.2926} \\
 & Extended & \textbf{0.0115} & 1.3103 \\
 & No Dataset Description & 0.0155 &  1.4226\\
 & Wrong Dataset Desc. & 0.0116 & 1.3952 \\
 & No Instruction & 0.8672 &  1.8801\\
\midrule
\multirow{5}{*}{DS-R1} & Concise & 1.2550 & 1.4059 \\
 & Extended & \textbf{0.9427} & 1.4059 \\
 & No Dataset Description & 1.1567 & 1.4059 \\
 & Wrong Dataset Desc. & 1.2477 & \textbf{1.3963} \\
 & No Instruction & 1.2762 & 1.4465 \\
\midrule
\multirow{5}{*}{LLaMA 3.3} & Concise & \textbf{0.0322} & 1.5336 \\
 & Extended & 0.0788 & 1.5204 \\
 & No Dataset Description & 0.0606 & 1.4933 \\
 & Wrong Dataset Desc.& 0.0596 & \textbf{1.4689} \\
 & No Instruction & 1.2734 & 1.6243 \\
\midrule
\multirow{5}{*}{GPT-4o} & Concise & 0.8100 & \textbf{1.2908} \\
 & Extended & \textbf{0.4187} & 1.3102 \\
 & No Dataset Description & 0.7989 & 1.3164 \\
 & Wrong Dataset Desc. & 0.7322 & 1.3157 \\
 & No Instruction & 1.0897 &  1.4007 \\
\bottomrule
\end{tabular}
\label{tab:RQ5_prompt_strategies}
\end{table}

\section{Discussion}
Taken together, the results across all research questions reveal a nuanced but constrained picture of current LLM capabilities in engineering-relevant tasks. 

LLMs show partial competence in handling transitional relations and local causal inferences (RQ2, RQ4), as well as in generating internally consistent world models under bounded conditions (RQ1). 
They can also extract functional intent and application context when these are directly implied by structural cues (RQ3), and exhibit modest success in forecasting simple, cyclic system behaviors (RQ5). 
However, performance deteriorates markedly as task complexity increases, particularly when success depends on resolving non-local dependencies, executing abstract generalization, engaging in formal modeling, or reasoning about dynamic systems characterized by feedback and non-linearity. 
These patterns collectively suggest that current LLMs are optimized for surface-level pattern recognition and localized reasoning, but fall short in structural comprehension, abstraction of formal constructs, and coherent causal modeling, which are essential capabilities for engineering-grade reliability.

Additionally, across nearly all evaluated dimensions, except the fith research question regarding the forecasting capabilities, the cloud-based GPT-4o model outperformed locally hosted counterparts. 
As the GPT-4o model is estimated to have about three times more parameters than the locally hosted models, this was expected. 
But, it underscores the potential value of enabling secure, privacy-preserving access to high-performance cloud-based LLMs in industrial settings. 

A further notable behavioral pattern concerns the models’ tendency to generate overextended or speculative responses rather than issuing a straightforward negation.
Instead of giving a simple no, LLMs tend to expand the use case and give more complex answers, leading to incorrect answers.
Similarly, models usually propose multiple candidate solutions rather than converging on a single, definitive answer, which introduces ambiguity in high-precision tasks. 
The evaluation shows that especially locally hosted LLMs currently work best as tools for human engineers, for example to support hypothesis generation, to narrow down potential failure modes or to investigate design alternatives. 
However, their current limitations preclude their autonomous use in challenging engineering scenarios. 
At present, they are not in a position to replace human expertise, but to supplement it under careful supervision.

\smallskip
What follows from this for the application of LLMs in engineering?
\noindent \emph{I. Abstraction:} Avoid tasks that require a high level of abstraction, e.g., the use of type-level abstractions or abstract concepts linked to the real world, such as intentions. The closer the task is to a concrete, instance-level problem, the better.
\noindent \emph{II. Models:} Avoid quantitative tasks; stick to symbolic, model-based tasks. If quantitative tasks such as time-series analysis or prognosis are needed, it is better for the LLM to generate a neural network or a simulation model, respectively.
\noindent \emph{III. Space:} Avoid problems that require far-reaching transitional relations between entities or complex causalities. Stick to local problems.
\noindent \emph{IV. Causality:} Forward reasoning (e.g., system optimization) works much better than backward reasoning (e.g., diagnosis). Avoid failure modes (e.g., FMEA); stick to normal operations.
\noindent \emph{V. LLMs:} The improvement of large LLMs compared to smaller ones is non-linear. LLMs are best used as assistant systems. If they are used as agents without supervision, avoid points I–IV above.

\section{Conclusion}
This paper has systematically evaluated the capabilities of LLMs in addressing engineering-related tasks, structured around five research questions that target core requirements of the engineering domain. For this analysis, the paper has introduced a new curated, publically available data set comprising forecasting tasks and more than 100 real-world engineering questions using two real-world systems. To the best of our knowledge, this is one of the first contributions using such a broad range of engineering tasks and such complex systems. 
The analysis of the research questions indicates that while LLMs demonstrate notable competence in localized, sequential, and pattern-based reasoning, their abilities remain constrained in several critical areas.

Specifically, LLMs perform surprisingly well in tasks involving the inference of system purpose from structural cues, the preservation of internal consistency across incremental updates, and the prediction of behaviors in simple, cyclic systems. Here especially, LLMs with a high number of latent variables perform very reliable for most engineering tasks---while smaller LLMs perform significantly worse. 

However, their performance deteriorates markedly when confronted with tasks that require non-local dependencies, type-level abstraction, formal causal reasoning, or the modeling of complex physical dynamics. These limitations underscore a fundamental misalignment between the statistical foundations of current LLMs and the structured, constraint-driven reasoning that characterizes engineering disciplines. 
Moreover, LLMs frequently tended to overgeneralized or speculative completions rather than issuing explicit rejections or selecting a single, well-supported solution.  This introduces ambiguity and compromises their reliability in engineering contexts, where precision and correctness are essential.

Not surprisingly given the nature of LLMs, they perform better for symbolic AI tasks such as (world) model maintenance, causal analysis or spatial analysis---and the larger the LLM the better they perform. For sub-symbolic (i.e., quantitative) tasks such as time-series analysis and prognosis, the performance deteriorates. So while general-purpose LLMs might replace engineers at some (symbolic) tasks, they will not replace other time-series-based AI systems, e.g., for anomaly detection.

Collectively, the results of this analysis suggest that while general-purpose LLMs are not yet suited for autonomous deployment in complex engineering tasks, they offer value as assistive tools for human experts, e.g., supporting hypothesis generation, failure mode analysis, and design exploration. 
Realizing their full potential in engineering workflows will require architectural adaptation, formal integration with domain models, and rigorous oversight to ensure interpretability, safety, and correctness in practice.

Looking ahead, we intend to advance this work by developing evaluation metrics that more explicitly capture structural, causal, and domain-specific correctness. Given the rapid evolution of the LLM landscape, periodic reevaluation with emerging models will be essential. We also plan to extend the scope of analysis by incorporating domain-specialized models and hybrid approaches that integrate formal reasoning or physics-based modeling, which may help to overcome current limitations.


\medskip
\emph{Acknowledgment: This research as part of the projects EKI, SmartShip and LaiLa is funded by dtec.bw – Digitalization and Technology Research Center of the Bundeswehr which we gratefully acknowledge. dtec.bw is funded by the European Union – NextGenerationEU.}

\bibliographystyle{abbrv}
\bibliography{bib}

\end{document}